\documentclass[11pt]{article}
\usepackage[T1]{fontenc}
\usepackage{lmodern}
\usepackage[margin=1in]{geometry}
\usepackage{authblk}
\usepackage{amssymb}
\usepackage{amsmath}
\usepackage[ruled,linesnumbered]{algorithm2e}
\usepackage{color,array,amsthm}
\usepackage{graphicx}
\usepackage{bm}
\usepackage{booktabs,tabularx,multirow,makecell}
\usepackage{tikz}
\usetikzlibrary{arrows.meta,positioning,fit}
\usepackage[hidelinks]{hyperref}
\usepackage{microtype}
\usepackage{placeins}

\begin{document}

\title{Task-Driven Three-Layer Distributed Scheduling for Emergency Earth Observation in Large Low-Earth-Orbit Constellations}

\author[1,2]{Qian Yin\thanks{\href{mailto:224201024@csu.edu.cn}{224201024@csu.edu.cn}}}

\author[2]{Xinwei Wang\thanks{Corresponding author: \href{mailto:xinwei.wang@qmul.ac.uk}{xinwei.wang@qmul.ac.uk}}}

\author[1]{Guohua Wu\thanks{Corresponding author: \href{mailto:guohuawu@csu.edu.cn}{guohuawu@csu.edu.cn}}}
\affil[1]{Central South University, Changsha, China}
\affil[2]{Queen Mary University of London, London, UK}
\date{}

\maketitle

\begin{abstract}
Large low-Earth-orbit (LEO) Earth-observation (EO) constellations offer frequent access to geographically dispersed ground targets, but emergency requests may arrive after committed routine-plan execution has begun.
The resulting dynamic emergency observation scheduling problem (DEOSP) requires urgent tasks to be inserted under intermittent ground contact without excessive routine-plan disruption.
To address DEOSP, we propose a task-driven three-layer distributed scheduling (T3L-DS) method, which represents task demand and sensor footprints on a common geographic grid and forms temporary clusters from observation capabilities and current inter-satellite links.
For intra-cluster coordination, T3L-DS introduces onboard dual-plan bidding and joint marginal evaluation.
It also designs an inter-cluster coordination mechanism for unresolved demand. 
Extensive computational experiments compare T3L-DS with centralised simulated annealing (SA), an adapted selective time-variant better reply process (A-SeTVBRP), and a conventional contract-net protocol (CNP). 
T3L-DS achieves the highest emergency coverage among the distributed methods, with average relative improvements of approximately 2.8\% and 17.1\% over A-SeTVBRP and CNP, respectively.
Its average relative gap from SA is approximately 7.1\%.
Under conflict-enhanced loads, it reduces routine-coverage loss by approximately 57.9\% and 87.7\% relative to A-SeTVBRP and CNP, respectively.
The ablation study confirms the contribution of the proposed coordination enhancements.
Overall, the results show that T3L-DS provides an effective distributed approach to DEOSP.
\end{abstract}

\medskip

\section{Introduction}
\label{sec:introduction}

Advances in spacecraft manufacturing and reusable launch services have accelerated the deployment of large low-Earth-orbit (LEO) constellations.
By 2026, public satellite population statistics listed more than 16,000 active satellites in Earth orbit.
More than 15,000 were in LEO, including about 1,400 payloads classified under imaging, radar imaging, Earth-observation (EO) science, or meteorological missions.\footnote{Satellite population and mission-category data are from Jonathan McDowell's public statistics: \url{https://planet4589.org/space/stats/active.html} and \url{https://planet4589.org/space/stats/omission.html}.}
For EO missions, large LEO constellations can provide frequent access and wide spatial coverage \cite{wang2020Agile,ferrari2024Satellite}.
These features are important for time-critical applications such as disaster assessment, maritime rescue, and rapid environmental monitoring.
In these applications, useful information must be collected within the valid observation window.

This motivates the dynamic emergency observation scheduling problem (DEOSP), in which emergency EO requests arrive during the execution of a committed routine observation plan \cite{wuDynamic2024,liu2022Bottom}.
At each arrival, every satellite has a current attitude and a sequence of completed or ongoing activities, protected future activities, and adjustable future activities.
Completed, ongoing, and protected activities remain fixed.
The scheduler then assigns emergency tasks within the adjustable future portion of the plan while limiting the loss of routine observations.

These scheduling decisions are difficult to centralise because constellation-wide coordination requires a current global execution state \cite{wuSurveyAutonomous2022,zilberstein2025Decentralized}.
Intermittent ground contacts delay the collection of satellite activity sequences and attitude states, and some information may already be outdated when a revised plan is ready \cite{duMultiSatelliteAutonomous2019}.
Detailed rescheduling is therefore better performed onboard by the satellites that can serve the emergency requests and have direct access to their current local states.
The ground centre retains event-level coordination by providing request data, predicted visibility, and topology information, while the relevant satellites construct and coordinate detailed schedule changes onboard.

Existing studies provide useful foundations, but they do not fully address the information and plan-modification requirements of DEOSP.
Centralised exact, heuristic, and metaheuristic methods provide strong schedules for planning-horizon problems \cite{pengExact2020,wuDivideConquer2022,zeng2026LongTerm}, but their execution-time use depends on timely global information and becomes costly as the candidate set grows.
The term distributed covers different decision settings in the current literature.
Krigman et al.~\cite{krigmanDSTS2024} distribute scheduling decisions among request-owning users rather than spacecraft, so their formulation does not define onboard plan modification from current satellite states.
Feng et al.~\cite{fengDistributed2023} and Zilberstein et al.~\cite{zilberstein2025Decentralized} develop distributed scheduling methods, but their planning still runs at a ground centre with access to constellation-wide information.
Yang et al.~\cite{yangDistributedSatellitesDynamic2025} propose a distributed potential-game method for allocating time-windowed observation grids, but assume real-time communication among satellites, which cannot be guaranteed in practice.
In practice, DEOSP  requires relevant satellites to modify their current onboard plans without a synchronised constellation-wide execution state and to coordinate through the links available at the time of each event.

To meet these requirements, this paper proposes a task-driven three-layer distributed scheduling (T3L-DS) framework.
Task-driven coordination rebuilds its scope for each emergency wave from the received requests, observable satellites, and current inter-satellite-link (ISL) topology.
It differs from constellation-wide schedule reconstruction \cite{wuDivideConquer2022,zeng2026LongTerm} and from resource-driven partitioning based on fixed groups or static ownership \cite{fengDistributed2023}.
In the T3L-DS, the ground layer maps requests and sensor footprints to a common geographic grid and forms temporary task-specific clusters.
The satellite layer performs local dual-plan bidding from its current onboard plan state after deterministic feasibility checks.
The cluster layer evaluates overlapping bids and reallocates tasks that remain unscheduled, without requiring a constellation-wide execution state.
This study adopts a hierarchical hexagonal geospatial indexing system (H3) as its geographic grid system \cite{h3docs}, while T3L-DS can also operate on conventional targets or predefined regional subtasks instead of grid cells.

In summary, this work makes three contributions:
\begin{itemize}
\item For the first time, we present a distributed formulation of DEOSP, using a common geographic-grid representation for point and area requests and satellite footprints while separating protected activities from adjustable future activities in the committed plan.

\item We develop T3L-DS to form an event-specific coordination scope and combine onboard dual-plan bidding, cluster-level marginal evaluation, and inter-cluster reallocation of unscheduled demand.

\item We conduct comparative and ablation experiments with mixed point and area requests under nominal and conflict-enhanced conditions.
The results show that T3L-DS provides the strongest emergency coverage among the distributed methods while limiting disturbance to the routine plan.
\end{itemize}

The rest of the paper is organised as follows.
Section~\ref{sec:related} reviews related work.
Section~\ref{sec:problem} describes DEOSP and presents its mathematical model.
Section~\ref{sec:t3ldsr_framework} describes the T3L-DS method.
Section~\ref{sec:experiments} presents the experimental design and analyses the results.
Section~\ref{sec:conclusion} concludes the paper.

\section{Related Work}
\label{sec:related}

Research related to DEOSP spans spatial representation, constellation scheduling, and distributed task allocation.
Point targets are commonly modelled as discrete observation opportunities, whereas wide or irregular regions are divided into subregions, grids, candidate observations, or attitude-dependent footprints.
Such representations support the joint allocation of coverage and observation time across several satellites \cite{jiMission2019a,xuMultisatellite2020}, while multi-objective formulations also consider image quality, resource use, and response time \cite{chenMultiobjective2020,chenLargescale2023}.
For agile and super-agile satellites, the manoeuvre path couples the spatial representation directly to observation time and coverage \cite{luMultiple2023,changMultistrip2023,wangVersatile2022}.

Adaptive subdivision and nested grids improve geometric fidelity \cite{xingAdaptive2024,heBalancing2020}, and resampling, multi-stage optimisation, and specialised heuristics limit the resulting search cost \cite{guLarge2022,chatterjeeMultistage2024,kandepiAgile2024}.
Grid-based formulations have also generated observation tasks or represented attitude-dependent coverage \cite{xuHeatGridASR2024,wangVersatile2022}.
In most of these studies, the grid primarily supports spatial discretisation or geometric search.
DEOSP requires one spatial identity to support demand decomposition, bidding, service credit, schedule-disturbance accounting, and residual-demand processing. The identity must remain consistent across satellites and scheduling stages to control duplicate credit and record partial completion \cite{wang2020Agile,ferrari2024Satellite}.

Centralised methods remain valuable benchmarks because they model constellation-wide interactions \cite{wang2016scheduling}.
Branch-and-bound, maximum-independent-set, and divide-and-conquer algorithms have addressed larger EO scheduling instances \cite{pengExact2020,eddyKochenderferMIS2021,wuDivideConquer2022}.
Other studies incorporate time-dependent value, repeated observations, decomposition, learning, or clustering while retaining a global planning model \cite{liIntervalProfit2024,wangHanLeusASR2025,qiDecomposeandlearn2025a,gallouaASR2025}.
A recent two-stage method also uses heuristic initialisation and global optimisation for long-term multi-region scheduling \cite{zeng2026LongTerm}.
Such methods provide useful quality references, but execution-time use requires the ground centre to collect a changing constellation state before it rebuilds the schedule.

Distributed formulations differ in both their decision variables and their coordination objectives.
Krigman et al. \cite{krigmanDSTS2024} formulate a distributed constraint optimisation problem in which request-owning users act as agents and exchange messages without first disclosing all requests to a central authority.
The distributed search protects request ownership, but it does not place schedule construction onboard the satellites.
Feng et al. \cite{fengDistributed2023} assign local schedule construction to satellite agents and resolve conflicts through iterative exchanges.
Zilberstein et al. \cite{zilberstein2025Decentralized} decompose the global scheduling problem into geometric neighbourhoods and apply decentralised stochastic search within those subproblems.
Both approaches begin from a defined request set and its observation opportunities, and their reported implementations retain constellation-level problem information.
Yang et al. \cite{yangDistributedSatellitesDynamic2025} take a different route by deriving satellite utilities from a global grid-allocation objective.
Their method passes a shared allocation file through sequential better replies and broadcasts the converged allocation under an assumption of real-time inter-satellite communication.
Research on autonomous EO systems has further established the value of onboard planning under limited ground contact \cite{wuSurveyAutonomous2022,duMultiSatelliteAutonomous2019}.
These studies focus mainly on creating a task allocation or a new schedule, rather than updating a partly executed routine plan after emergency requests arrive.

The contract-net protocol (CNP) offers a natural manager-contractor structure for such local decisions \cite{smith1980The}, and its variants allow heterogeneous EO resources to evaluate opportunities under local constraints \cite{liu2022Bottom}.
A standard contract nevertheless treats a task or bid as an indivisible unit.
Area-target bids may overlap on only some cells, so whole-bid acceptance can duplicate credit and whole-bid rejection can discard useful coverage.
Learning-based schedulers can supply local preferences quickly after training \cite{herrmannSchaubRL2023,herrmannSingle2024,songRLGA2023}, but feasibility and cross-satellite consistency still require explicit coordination rules.
Robust EO scheduling protects plans against uncertainty before execution \cite{wang2019Robust}. 
DEOSP instead concerns schedule modification after emergency demand becomes known during routine-plan execution.
For the distributed solution developed in this study, satellites that can observe the current emergency demand are organised into temporary clusters using task visibility and available ISLs.
Each cluster provides a local coordination scope, while unresolved demand can be transferred to a neighbouring cluster.

\begin{figure*}[t]
\centering
\includegraphics[width=0.96\textwidth]{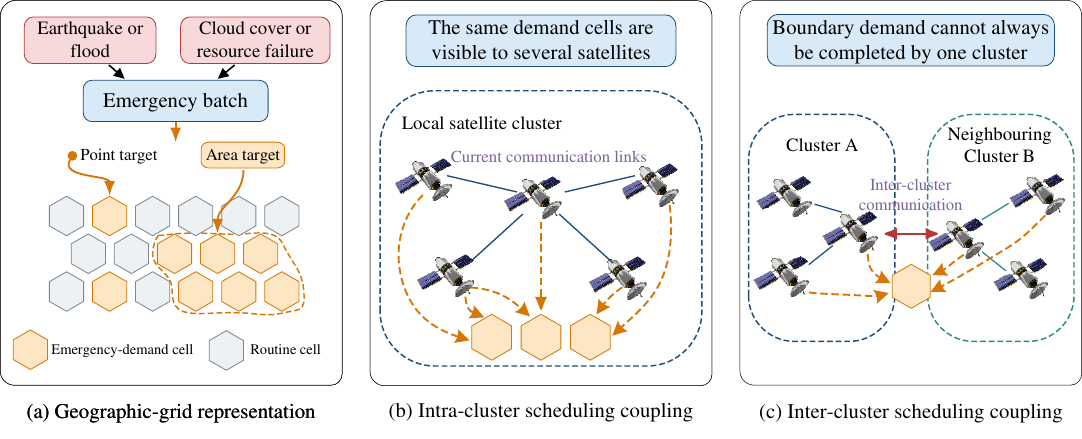}
\caption{Research setting of DEOSP.
The figure shows geographic-grid conversion, intra-cluster coupling, and inter-cluster coupling.}
\label{fig:problem_setting_example}
\end{figure*}

In a nutshell, existing studies have not established a distributed formulation of DEOSP for constellations whose current plans are held locally and whose ISLs vary during execution. In this setting, an emergency insertion changes both the adjustable routine plan of one satellite and the demand still available to the others. T3L-DS addresses this problem through distributed onboard scheduling and coordination within and between temporary clusters over the available ISLs.

\section{Problem Description and Modelling}
\label{sec:problem}

In this section, we first describe the problem operational setting and the distributed information boundary, and
then introduce a common geographic-grid representation for emergency demand and satellite observation footprints. In the end, we introduce a centralised reference model that clarifies the objective, constraints, and schedule-disturbance terms used by the distributed method.

\subsection{Problem Description of DEOSP}
\label{sec:workflow}

During routine operation, a large LEO EO constellation follows a committed plan with prescribed execution times and attitudes.
Emergency requests caused by disasters, cloud cover, or resource failures may arrive while this plan is being executed \cite{wuDynamic2024,liu2022Bottom}.
Each request specifies a point or area target, a release time, a priority, and an observation window within which the target must be observed.
Following the unified grid characterisation in \cite{yinHeatmapGrid2026}, a common geographic grid converts both request types and satellite footprints to the same cell representation, and each grid-cell-target pair forms an atomic demand unit.
At each arrival, DEOSP assigns these units using predicted observation opportunities and the current local plans and attitudes of the satellites, which the ground planning centre may not hold in a fully synchronised form because contacts are intermittent.
The resulting schedule changes must satisfy request windows, visibility intervals, and attitude-transition requirements.
They may affect only adjustable future activities; completed, ongoing, and protected activities remain fixed.
DEOSP seeks to maximise emergency coverage within the request windows while limiting the loss of valid routine coverage, without reconstructing the full mission plan.
To support distributed decisions under this information boundary, T3L-DS forms temporary clusters according to the current demand-satellite visibility relation and available ISLs.
Each cluster coordinates its assigned demand locally, and unresolved demand may be transferred to a neighbouring cluster.

The example in Figure~\ref{fig:problem_setting_example} considers an emergency wave arriving at time $t_e$ with point and area requests while cloud cover has made part of the routine plan ineffective.
Grid conversion gives the requests and satellite footprints a common cell identity.
In one region, several satellites can observe overlapping cells, but their committed activities, attitudes, and adjustable time intervals differ.
Their local decisions are coupled through competition for cell credit and through the different routine activities disturbed by an insertion.
Some boundary cells may remain unresolved after this intra-cluster coordination, even though a neighbouring cluster still has suitable visibility and communication access.
Inter-cluster coordination must transfer these cells without losing their identity or creating duplicate credit.

\subsection{Notation and assumptions}

Table~\ref{tab:notation} links the symbols used in the centralised reference model to those used later for distributed coordination. 
The request and execution-state notation is introduced below, while the cluster and candidate symbols are used in Section~\ref{sec:t3ldsr_framework}.

\begin{table}[htb]
\centering
\caption{Main symbols and notation}
\label{tab:notation}
\scriptsize
\setlength{\tabcolsep}{3pt}
\renewcommand{\arraystretch}{1.02}
\begin{tabularx}{\columnwidth}{>{\centering\arraybackslash}p{0.39\columnwidth}X}
\toprule
Symbol & Definition \\
\midrule
$\mathcal{S},N$ & Satellite set and number of satellites. \\
$\mathcal{W},e,t_e$ & Emergency waves set, wave index, and arrival time. \\
$\mathcal{R}^{E}_e$ & Requests released in wave $e$. \\
$P_r,t_r^{\mathrm{rel}},t_r^{\mathrm{ddl}},\omega_r$ & Region, time window, and priority of request $r$. \\
$\mathcal{G}_{\ell},C_g$ & Grid cells at resolution $\ell$ and polygon of cell $g$. \\
$\mathcal{G}_r,\mathcal{D}^{E}_e$ & Grid representation and atomic emergency demands. \\
$\sigma_s^e,\bar\Pi_s^e$ & Execution state and committed sequence of satellite $s$. \\
\makecell{$\bar\Pi_{\mathrm{exe}}^e,\bar\Pi_{\mathrm{fix}}^e,$\\$\bar\Pi_{\mathrm{adj}}^e$} & Executed, protected, and adjustable activities. \\
$\mathcal{P}_s^e,p$ & Candidate observation of satellite $s$ and one candidate. \\
$F_p,\Gamma_p,\Omega_e$ & Candidate footprint, covered cells, and feasible service pairs. \\
$\mathcal{E}_s^e$ & Pairwise candidate incompatibilities. \\
$x_p^e,z_{p,q}^e,v_g^e$ & candidate selection, demand service, and routine-cell invalidation variables. \\
$\mathcal{O}_s^e,\Gamma_o,\mathcal{S}^{\mathrm{vis}}_q$ & Future opportunities, opportunity footprints, and visible satellites. \\
$\mathcal{G}_{\mathrm{ISL}}^e,\mathcal{B}^e,K_{\max}$ & ISL graph, temporary clusters, and size limit. \\
$\chi,\mathcal{X}_s^e,\mathcal{Y}(\chi)$ & Local candidate, candidate set, and covered cell-target pairs. \\
$\Pi_s^{A},\Pi_s^{B}$ & Primary and diverse local plans. \\
$\mathcal{L}_b,\Delta J(\chi\mid\mathcal{L}_b)$ & Credited pairs and marginal gain in cluster $b$. \\
$\mathcal{D}_{\mathrm{res}}$ & Residual-demand set. \\
\bottomrule
\end{tabularx}
\end{table}

Let $\mathcal{S}=\{s_1,\ldots,s_N\}$ denote the satellite set, and let $\mathcal{T}=[0,T]$ denote the mission horizon.
Emergency requests arrive in waves indexed by $e\in\mathcal{W}$.
Wave $e$ arrives at time $t_e$ and contains the request set $\mathcal{R}^{E}_e$.
Each request $r$ has a geographic region $P_r$, a release time $t_r^{\mathrm{rel}}$, a deadline $t_r^{\mathrm{ddl}}$, and a priority $\omega_r$.
The interval $[t_r^{\mathrm{rel}},t_r^{\mathrm{ddl}}]$ defines the observation window of request $r$.

At the arrival time $t_e$, each satellite $s$ is characterised by an execution state
\begin{equation}
\sigma_s^e=(\bar\Pi_s^e,\bm\theta_s^e,t_s^{\mathrm{ava}}).
\label{eq:execution_state}
\end{equation}
Here, $\bar\Pi_s^e$ is the committed activity sequence stored onboard satellite $s$, and $\bm\theta_s^e$ is its current attitude.
The term $t_s^{\mathrm{ava}}$ denotes the earliest time at which the schedule can be changed.
The committed constellation schedule $\bar\Pi^e=\bigcup_{s\in\mathcal{S}}\bar\Pi_s^e$ is an external input to DEOSP.
T3L-DS does not create or independently optimise that schedule.
For modelling purposes, the committed schedule is partitioned as
\begin{equation}
\bar\Pi^e=
\bar\Pi_{\mathrm{exe}}^e\mathbin{\dot\cup}
\bar\Pi_{\mathrm{fix}}^e\mathbin{\dot\cup}
\bar\Pi_{\mathrm{adj}}^e.
\label{eq:schedule_partition}
\end{equation}
The set $\bar\Pi_{\mathrm{exe}}^e$ contains activities that have finished or are already being executed.
The set $\bar\Pi_{\mathrm{fix}}^e$ contains future activities protected by mission priority, operational commitment, or an uploaded execution segment.
Neither set can be changed after wave $e$ arrives.
Only the future activities in $\bar\Pi_{\mathrm{adj}}^e$ are available for insertion, pre-emption, or rescheduling.

The formulation is based on four operational assumptions.
First, each satellite has the onboard computing, inter-satellite communication, and target-recognition capability required to process an emergency request.
Second, emergency requests arrive as event-triggered batches.
Requests in the same batch share a release event, but they may have different observation windows and priorities \cite{wuDynamic2024,liu2022Bottom}.
Third, the ground segment cannot communicate with every satellite in real time.
It sends compact requests and coordination information to relevant satellites, rather than collecting every detailed onboard schedule \cite{wuSurveyAutonomous2022,duMultiSatelliteAutonomous2019}.
Fourth, each grid-cell-target pair receives at most one credited service in one allocation attempt \cite{krigmanDSTS2024,yangDistributedSatellitesDynamic2025}.

\subsection{Geographic Grid-Based Demand and Resource Representation}
\label{sec:grid_representation}

Geographic gridding converts the continuous Earth surface into discrete cells with unique spatial indices.
When the grid is hierarchical, it also provides parent-child relations between resolutions.
These properties allow the scheduler to use the same spatial unit for demand decomposition, footprint coverage, duplicate removal, and residual-demand tracking.
Hexagonal cells have six adjacent directions over most of the grid, which is useful for compact regional representation and local spatial search \cite{yinHeatmapGrid2026,sahrLocation2008}.

Let $\mathcal{G}_{\ell}$ denote the cells of a geographic grid at resolution $\ell$, and let $C_g$ be the geographic polygon of cell $g\in\mathcal{G}_{\ell}$.
Note that the three-layer architecture can use other atomic spatial units, such as conventional targets or predefined regional subtasks, without changing the division of coordination responsibilities.
In this study, H3 is used to instantiate the geographic-grid representation because it provides hierarchical global indexing and direct cell-neighbour operations \cite{h3docs}.
The scheduling mechanisms require only an identifiable spatial unit, its coverage membership, and, where candidate grouping is used, its neighbourhood relation.

The cell representation of emergency request $r$ is
\begin{equation}
    \mathcal{G}_r =
    \left\{g\in\mathcal{G}_{\ell}: C_g\cap P_r\neq\emptyset\right\}.
    \label{eq:request_cells}
\end{equation}
A point request is mapped to the cell that contains its coordinate.
An area request is represented by the cells that intersect its polygon.
The atomic emergency-demand set is
\begin{equation}
    \mathcal{D}^{E}_{r}=\{q=(r,g):g\in\mathcal{G}_r\},
    \qquad
    \mathcal{D}^{E}_{e}=\bigcup_{r\in\mathcal{R}^{E}_e}\mathcal{D}^{E}_{r}.
\label{eq:atomic_demands}
\end{equation}
For $q=(r,g)$, let $r(q)=r$ and $g(q)=g$ denote its request and grid-cell components.

A candidate observation $p$ belongs to satellite $s(p)$.
It is described by an inserted observation interval $(t_p^{\mathrm{st}},t_p^{\mathrm{end}})$, an attitude state $\bm{\theta}_p$, and a ground footprint $F_p$.
It also has a manoeuvre or change cost $c_p^{\mathrm{chg}}$.
Its grid footprint is
\begin{equation}
    \Gamma_p =
    \left\{g\in\mathcal{G}_{\ell}: C_g\cap F_p\neq\emptyset\right\}.
    \label{eq:pattern_cells}
\end{equation}
Candidate $p$ can serve demand $q=(r,g)$ only when $g$ belongs to $\Gamma_p$, the inserted observation starts no earlier than $t_{r(q)}^{\mathrm{rel}}$ and ends no later than $t_{r(q)}^{\mathrm{ddl}}$, and the sensor mode matches the request.
The set $\Omega_e$ contains all service pairs that satisfy these conditions.

The service value of a feasible pair is represented by the priority of its original request.
This gives the model a simple emergency-service term while keeping response-time feasibility inside the definition of $\Omega_e$.

\begin{figure}[ht]
\centering
\includegraphics[width=0.98\columnwidth]{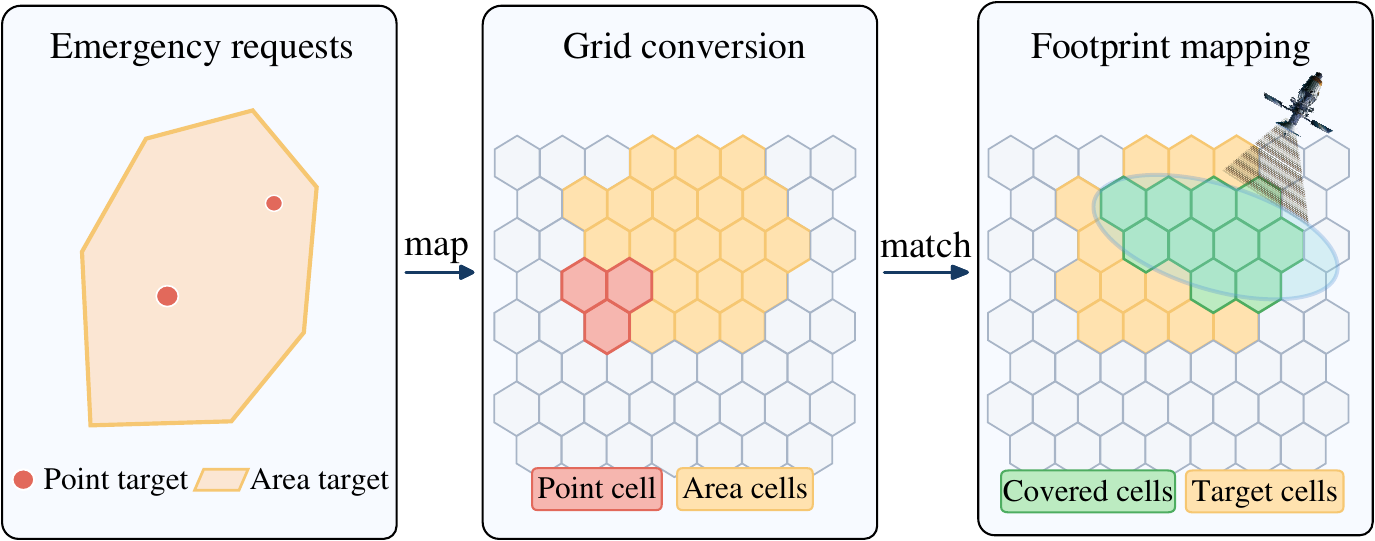}
\caption{Geographic-grid representation for emergency demand and observation footprints.
The figure shows how point requests, area requests, and satellite footprints are mapped to the same cell set.}
\label{fig:grid_mapping}
\end{figure}

Figure~\ref{fig:grid_mapping} illustrates the common spatial representation used by the model.
After the conversion, both request geometry and satellite-footprint geometry are represented by cell sets.
The model compares $\mathcal{G}_r$ and $\Gamma_p$ through set operations.
The intersection $\mathcal{G}_r\cap\Gamma_p$ gives the cells that candidate $p$ can add to request $r$.
The same cell identity is later used for service credit, duplicate removal, residual demand, and disturbance measurement.

\subsection{Dynamic Emergency Observation Scheduling Model}
\label{sec:repair_model}

For satellite $s$, two candidate observations $p,p'\in\mathcal{P}_s^e$ can conflict either because their observation intervals overlap or because the available transition time is shorter than the required manoeuvre time $\Delta_s(\bm{\theta}_p,\bm{\theta}_{p'})$.
For notation, let $p$ precede $p'$ when $t_p^{\mathrm{st}}\le t_{p'}^{\mathrm{st}}$.
The conflict set is
\begin{equation}
\begin{aligned}
\mathcal{E}_s^e=\{(p,p'):\;&p\neq p',\; s(p)=s(p')=s,\\
&t_p^{\mathrm{st}}\le t_{p'}^{\mathrm{st}},\\
&[t_p^{\mathrm{st}},t_p^{\mathrm{end}}]\cap
[t_{p'}^{\mathrm{st}},t_{p'}^{\mathrm{end}}]\neq\emptyset\\
&\text{or }t_{p'}^{\mathrm{st}}-t_p^{\mathrm{end}}
<\Delta_s(\bm{\theta}_p,\bm{\theta}_{p'})\}.
\end{aligned}
\label{eq:conflict_set}
\end{equation}

The set $\bar\Pi^e$ is divided into executed activities $\bar\Pi_{\mathrm{exe}}^e$, protected activities $\bar\Pi_{\mathrm{fix}}^e$, and adjustable future activities $\bar\Pi_{\mathrm{adj}}^e$.
The first two subsets cannot be changed by the rescheduling process.
The set $\bar\Pi_{\mathrm{fix}}^e$ contains routine activities that are still in the future but are protected by priority, operational commitment, or the start of an already uploaded execution segment.
Only activities in $\bar\Pi_{\mathrm{adj}}^e$ may be displaced by emergency insertions.
When a selected emergency candidate $p$ conflicts with an adjustable routine activity, the time-local cell-level update uses $t_p^{\mathrm{st}}$ as its boundary. Routine cell records that start before this boundary remain valid, whereas records that start at or after it become invalid.

Let $x_p^e\in\{0,1\}$ indicate whether the model selects candidate $p\in\mathcal{P}^e$.
Let $z_{p,q}^e\in\{0,1\}$ indicate whether candidate $p$ serves emergency demand $q$.
Let $v_g^e\in\{0,1\}$ indicate whether a selected candidate invalidates future routine cell $g$.
Here, $\mathcal{P}^e=\bigcup_{s\in\mathcal{S}}\mathcal{P}_s^e$.
For each candidate $p$, the set $\mathcal{Q}(p)$ contains only the adjustable future routine cells at or after the start time of $p$ that it can invalidate.
The centralised formulation is
\begin{equation}
\begin{aligned}
\max_{x^e,z^e,v^e}\quad &
\sum_{q\in\mathcal{D}^{E}_e}
\sum_{p:(p,q)\in\Omega_e}\omega_{r(q)}z_{p,q}^{e}\\
&-\lambda_{\mathrm{dis}}\sum_{g}c_g^{\mathrm{dis}}v_g^{e}
-\lambda_{\mathrm{chg}}\sum_{p}c_p^{\mathrm{chg}}x_p^e,
\end{aligned}
\label{eq:repair_obj}
\end{equation}
where $c_g^{\mathrm{dis}}$ is the disturbance cost of routine cell $g$, while $\lambda_{\mathrm{dis}}$ and $\lambda_{\mathrm{chg}}$ control the penalties for routine-plan disturbance and schedule change.
The objective is subject to
\begin{align}
&z_{p,q}^{e}\le x_p^{e}, &&\forall(p,q)\in\Omega_e,
\label{eq:repair_link}\\
&\sum_{p:(p,q)\in\Omega_e}z_{p,q}^{e}\le1,
&&\forall q\in\mathcal{D}^{E}_e,
\label{eq:repair_unique}\\
&x_p^{e}+x_{p'}^{e}\le1,
&&\forall(p,p')\in\mathcal{E}_s^e,\;s\in\mathcal{S},
\label{eq:repair_conflict}\\
&v_g^e\ge x_p^e,
&&\forall p\in\mathcal{P}^e,\;g\in\mathcal{Q}(p).
\label{eq:disruption_link}
\end{align}

\begin{figure*}[htb]
\centering
\includegraphics[width=0.96\textwidth]{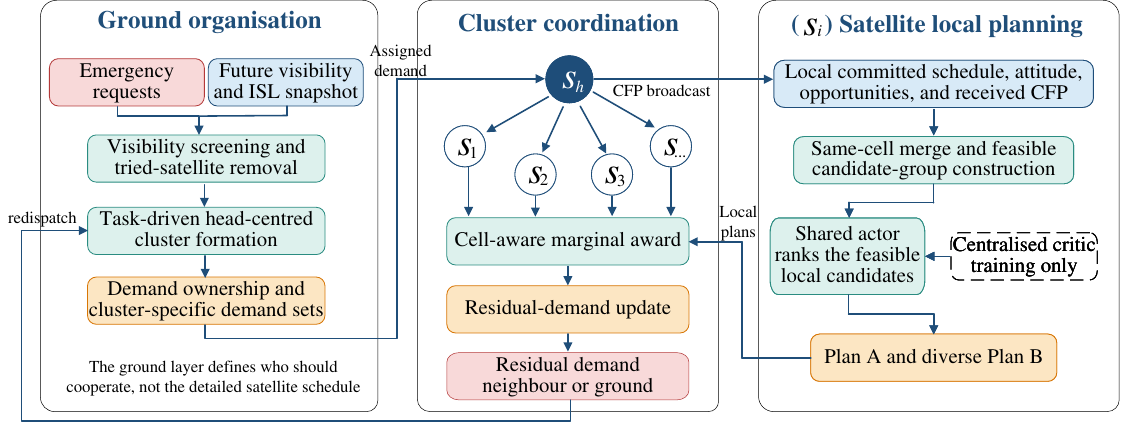}
\caption{Overall information flow of T3L-DS.
The ground segment uses request data, future visibility, and the ISL snapshot to define event-related coordination scopes.
The cluster layer sends calls for proposals (CFPs), and the satellites return local plans for cluster-level coordination.
Here, $s_h$ denotes the cluster head, and $s_i$ denotes a member satellite.}
\label{fig:architecture}
\end{figure*}

The model removes candidates that conflict with protected activities before optimisation.
Constraint~\eqref{eq:repair_link} links demand service to candidate selection.
Constraint~\eqref{eq:repair_unique} gives each emergency demand unit at most one credited service.
Constraint~\eqref{eq:repair_conflict} prevents overlapping or attitude-infeasible candidates on the same satellite.
Constraint~\eqref{eq:disruption_link} records which adjustable routine cells are invalidated by selected emergency insertions.
Deadline feasibility is handled before optimisation by constructing $\Omega_e$ only from candidate-demand pairs that finish within the request window.
Pairwise candidate incompatibilities define an incompatibility graph, in which maximum-value observation selection can be formulated as a maximum-weight independent-set problem \cite{eddyKochenderferMIS2021}.
This class of EO satellite scheduling problems is NP-hard \cite{wuFrequentPattern2023}.
Pre-emption rules and schedule-change costs extend this structure to the execution-time setting.
T3L-DS uses the formulation to define a common objective and feasibility boundary, while operating on the constellation state available when each emergency wave arrives.

\section{T3L-DS Method}
\label{sec:t3ldsr_framework}

\subsection{Overall framework}
\label{sec:framework_overview}
T3L-DS takes the committed schedule, current onboard states, future observation opportunities, and the current ISL graph as its event input.
For emergency wave $e$, $\mathcal{G}_{\mathrm{ISL}}^e=(\mathcal{S},\mathcal{Z}^e)$ is a link-availability snapshot generated from the predicted satellite positions and the adopted line-of-sight condition at the wave-arrival epoch, where $(s_i,s_j)\in\mathcal{Z}^e$ denotes an available direct ISL between satellites $s_i$ and $s_j$ \cite{wang2019onboard}
It assigns event-level organisation to the ground layer, overlapping allocation to temporary clusters, and detailed schedule changes to individual satellites.

Figure~\ref{fig:architecture} shows the information exchanged across the three layers.
The ground segment supplies request, visibility, and topology information; cluster heads issue CFPs and resolve overlapping bids; satellites retain their committed sequences and return compact local plans.
The head shown in Figure~\ref{fig:architecture} is selected during the event-specific cluster construction and is not a permanent role assigned in advance.
Only the cluster layer assigns final cell ownership, so each satellite can generate its local plan without access to a fully synchronised constellation-wide schedule.

\subsection{Task-driven temporary clustering}
\label{sec:clustering}

The ground segment identifies satellites with an opportunity that intersects both the cell and the request window:

\begin{equation}
\begin{aligned}
\mathcal{S}^{\mathrm{vis}}_q=\{s\in\mathcal{S}:\exists o\in\mathcal{O}_s^e,
&\ g(q)\in\Gamma_o,\\
&\ t_o^{\mathrm{end}}>t_{r(q)}^{\mathrm{rel}},\
t_o^{\mathrm{st}}<t_{r(q)}^{\mathrm{ddl}}\}.
\end{aligned}
\label{eq:visible_satellite_set}
\end{equation}
Here, $\Gamma_o$ denotes the grid-cell footprint of opportunity $o$.
Satellites recorded as having tried $q$ are excluded during redispatch.
A demand is declared infeasible only when no future opportunity remains; otherwise it stays pending if the current organisation cannot accept it.

Let $\mathcal{B}^e$ denote the temporary-cluster set for wave $e$, and let $b\in\mathcal{B}^e$ index one cluster.
Each currently unassigned satellite is first treated as a candidate cluster head.
Its directly linked unassigned neighbours are ranked by the number of current demands that they can observe, and at most $K_{\max}$ satellites are retained.
The candidate head and its retained neighbours form a head-centred cluster with a star topology, because every member has a direct ISL to the head, whereas links between members are not required.
The candidate cluster covering the largest number of distinct current demands is accepted, and its central satellite becomes the cluster head.
The satellites in the accepted cluster are then removed from the unassigned set, and the procedure continues until no further cluster can be formed.
Each demand is then assigned to the cluster containing the most satellites in $\mathcal{S}^{\mathrm{vis}}_q$, with current cluster load used to break ties.
Clusters are rebuilt at every wave, so this organisation follows the event rather than a fixed constellation partition.

\subsection{Satellite-local candidate generation and dual-plan construction}
\label{sec:local_planning}

After receiving a CFP, satellite $s$ first merges demand already served by a same-cell observation inside the emergency window.
It builds $\mathcal{X}_s^e$ for the remaining demand from future local opportunities.
A local candidate $\chi$ is the distributed representation of a candidate modification $p$ in the reference model.
Each candidate $\chi$ records one execution interval, one attitude, the cell-target set $\mathcal{Y}(\chi)$, and any lower-priority activities approved for pre-emption.
Completed cell-target records remove already served residual or routine demand before ranking; emergency demand may still require a new observation of the same cell outside the relevant emergency window.

Let $\mathcal{X}_{s,\mathrm{win}}^e$, $\mathcal{X}_{s,\mathrm{sep}}^e$, and $\mathcal{X}_{s,\mathrm{prio}}^e$ denote the candidates that satisfy task windows, time-attitude separation, and pre-emption priority, respectively.
The local feasibility rule is
\begin{equation}
\mathbb{I}^{\mathrm{feas}}_s(\chi)=
\mathbb{I}\!\left[
\chi\in
\mathcal{X}_{s,\mathrm{win}}^e\cap
\mathcal{X}_{s,\mathrm{sep}}^e\cap
\mathcal{X}_{s,\mathrm{prio}}^e
\right].
\label{eq:local_feasibility}
\end{equation}
The separation test uses $\Delta_s(\bm\theta,\bm\theta')$ from Eq.~\eqref{eq:conflict_set}.
Emergency candidates may pre-empt lower-priority routine activities but not committed emergency observations.
The implementation ranks only candidates that satisfy Eq.~\eqref{eq:local_feasibility}; an offline-trained local policy supplies this ordering, while all feasibility decisions remain deterministic.

\begin{figure}[!htb]
\centering
\includegraphics[width=0.65\columnwidth]{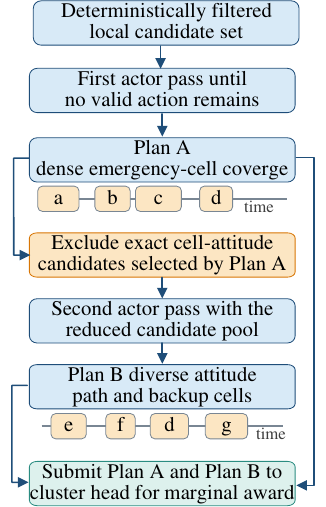}
\caption{Dual-plan construction from a deterministically filtered local candidate set.}
\label{fig:dual_plan_generation}
\end{figure}

As shown in Figure~\ref{fig:dual_plan_generation}, sequential selection with a local state update produces Plan~A.
The exact cell-attitude candidates used by Plan~A are then removed, and the same construction on the reduced set produces Plan~B.
Both plans carry their cell-target pairs, execution data, and approved pre-emption records to the cluster head.

\subsection{Intra-cluster contract-net coordination}
\label{sec:contractnet}

\begin{figure}[!htb]
\centering
\includegraphics[width=0.80\columnwidth]{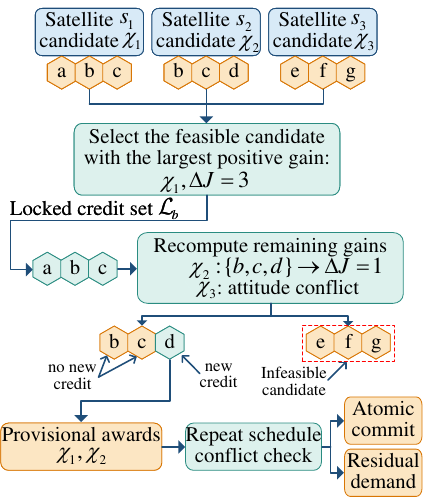}
\caption{Cell-aware marginal awards within one temporary cluster.
The locked set $\mathcal{L}_b$ stores credited cell-target pairs, and $\Delta J$ is recomputed after each provisional award.}
\label{fig:contract_net_detail}
\end{figure}

The cluster head expands the submitted plans into a candidate set $\mathcal{X}_b$ and evaluates candidates separately.
Let $\mathcal{L}_b$ contain the cell-target pairs already credited in cluster $b$.
The marginal gain of candidate $\chi$ is
\begin{equation}
\Delta J(\chi\mid\mathcal{L}_b)=
\left|\mathcal{Y}(\chi)\setminus\mathcal{L}_b\right|.
\label{eq:marginal_candidate_score}
\end{equation}
A candidate must also remain time-attitude compatible with earlier awards on the same satellite.
The head repeatedly selects the feasible candidate with the largest positive gain, adds only its new pairs to $\mathcal{L}_b$, and recomputes the remaining gains.
Hence, a partially overlapping bid can retain useful cells, while an already credited pair cannot receive duplicate credit.
Figure~\ref{fig:contract_net_detail} traces this update within one cluster. Each award extends the locked cell-target set, changes the marginal gains of overlapping candidates, and removes attitude-infeasible alternatives before the next selection.

\begin{algorithm}[htb]
\caption{Cell-aware marginal award and atomic commitment}
\label{alg:marginal_award}
\KwIn{Cluster demand set $\mathcal{D}_b$, candidate set $\mathcal{X}_b$ and current satellite schedules}
\KwOut{Committed set $\mathcal{C}_b$ and residual set $\mathcal{D}_{\mathrm{res}}$}
$\mathcal{L}_b,\mathcal{A}_b,\mathcal{C}_b,\mathcal{D}_{\mathrm{res}}\leftarrow\emptyset,\emptyset,\emptyset,\emptyset$\;
\While{$\mathcal{X}_b\neq\emptyset$}{
  $\mathcal{F}_b\leftarrow\{\chi\in\mathcal{X}_b:\operatorname{Compatible}(\chi,\mathcal{A}_b)\}$\;
  \If{$\mathcal{F}_b=\emptyset$}{break\;}
  $\chi^*\leftarrow\arg\max_{\chi\in\mathcal{F}_b}\Delta J(\chi\mid\mathcal{L}_b)$\;
  \If{$\Delta J(\chi^*\mid\mathcal{L}_b)=0$}{break\;}
  $\mathcal{A}_b\leftarrow\mathcal{A}_b\cup\{\chi^*\}$,
  $\mathcal{L}_b\leftarrow\mathcal{L}_b\cup\mathcal{Y}(\chi^*)$\;
  $\mathcal{X}_b\leftarrow\mathcal{X}_b\setminus\{\chi^*\}$\;
}
\ForEach{$\chi\in\mathcal{A}_b$}{
  \eIf{$\operatorname{CommitFeasible}(\chi)=1$}{
    $\operatorname{AtomicCommit}(\chi)$; $\mathcal{C}_b\leftarrow\mathcal{C}_b\cup\{\chi\}$\;
  }{
    $\mathcal{D}_{\mathrm{res}}\leftarrow\mathcal{D}_{\mathrm{res}}\cup\mathcal{Y}(\chi)$\;
  }
}
{$\mathcal{D}_{\mathrm{served}}\leftarrow\bigcup_{\chi\in\mathcal{C}_b}\mathcal{Y}(\chi)$}\;
{$\mathcal{D}_{\mathrm{res}}\leftarrow\mathcal{D}_{\mathrm{res}}\cup(\mathcal{D}_b\setminus\mathcal{D}_{\mathrm{served}})$}\;
\end{algorithm}

Algorithm~\ref{alg:marginal_award} separates provisional allocation $\mathcal{A}_b$ from committed observations $\mathcal{C}_b$.
The final check uses the latest satellite schedule.
An accepted candidate and its approved pre-emptions are committed atomically; a new conflict instead returns the affected demand to $\mathcal{D}_{\mathrm{res}}$.

\subsection{Inter-cluster coordination and event termination} \label{sec:intercluster}

The residual set produced by Algorithm \ref{alg:marginal_award} contains uncredited emergency demand and demand returned by failed atomic commitments.
The current managing cluster first conducts a further intra-cluster allocation attempt using the remaining feasible candidates.
An unresolved demand unit may then be forwarded to a neighbouring cluster if the two cluster heads have a direct ISL in the current topology.
If the inter-cluster attempt does not yield an assignment, the ground segment updates the demand-satellite visibility relation, rebuilds the event-specific clusters under the latest ISL topology, and redispatches the demand unit.
At every stage, exactly one cluster retains allocation authority for each active demand unit, preventing concurrent or duplicate awards.

Routine cell demands created by an approved pre-emption are subject to a more restrictive rescheduling rule.
They may undergo one local rescheduling attempt followed, if necessary, by one neighbouring-cluster attempt, but they are not eligible for subsequent ground-level redispatch.
An unresolved emergency demand unit is deferred while a feasible future opportunity remains and is classified as infeasible only when no such opportunity exists.
The coordination process for wave $e$ terminates when every demand unit has been served, classified as infeasible, deferred, or removed after exhausting its permitted one-pass rescheduling path.
The associated cluster configuration is not retained for subsequent emergency waves.

\section{Computational Experiments}
\label{sec:experiments}

\subsection{Experimental protocol and scenario design}

Each experiment begins with a committed routine schedule generated by a cost-effective lazy forward (CELF) greedy procedure.
All methods treat this schedule as an external input and may modify only its adjustable future activities.
The event-driven simulator releases emergency point and area requests in successive waves during plan execution.
Within each instance, all methods receive identical committed schedules, emergency requests, observation opportunities, and satellite states.
Based on preliminary experiments, the penalty coefficients in the reference model were set to $\lambda_{\mathrm{dis}}=10$ and $\lambda_{\mathrm{chg}}=1$ for all computational experiments.

The experiments use H3 resolution~6, to which all reported grid-cell counts refer.
The orbital data used in this study are taken from EOS-Bench, a comprehensive benchmark for Earth observation satellite scheduling \cite{YinComprehensive_2026}.
The benchmark's 500-satellite constellation comprises 25 orbital planes with 20 satellites per plane and is generated from a seed orbit with semi-major axis 7013.62362~km, eccentricity 0.000898, inclination $98.04^\circ$, right ascension of the ascending node $57.345^\circ$, argument of perigee $101.516^\circ$, and true anomaly $96.459^\circ$.
Smaller constellations are uniform subsets of this configuration.
All experiments were conducted in Python 3.10 on Windows 11 using an Intel Core i7-10700K CPU at 3.80 GHz, 64 GB RAM, and an NVIDIA GeForce RTX 3090 GPU.

\begin{table}[htb]
\centering
\caption{Experimental scenario configurations}
\label{tab:scenarios}
\fontsize{6.3pt}{7.1pt}\selectfont
\setlength{\tabcolsep}{1.2pt}
\renewcommand{\arraystretch}{0.98}
\begin{tabular*}{\columnwidth}{@{\extracolsep{\fill}}ccccccccc@{}}
\toprule
\multirow{2}{*}{Case} & \multirow{2}{*}{Satellites} &
\multicolumn{2}{c}{Routine targets} &
\multicolumn{2}{c}{Emergency targets} &
\multirow{2}{*}{Waves} &
\multirow{2}{*}{{\makecell{Emergency\\window (min)}}} &
\multirow{2}{*}{{\makecell{Mean emergency\\H3 cells}}}\\
\cmidrule(lr){3-4}\cmidrule(lr){5-6}
& & Point & Region & Point & Region & & & \\
\midrule
A1 & 500 & 8000 & 3000 & 50 & 25 & 5 & 60-120 & 550\\
A2 & 500 & 8000 & 3000 & 100 & 50 & 10 & 60-120 & 1100\\
A3 & 500 & 8000 & 3000 & 300 & 150 & 15 & 60-120 & 3300\\
A4 & 500 & 8000 & 3000 & 400 & 200 & 20 & 60-120 & 4400\\
\addlinespace[0.8pt]
B1 & 500 & 4000 & 1500 & 200 & 100 & 15 & 60-120 & 2200\\
B2 & 500 & 8000 & 3000 & 200 & 100 & 15 & 60-120 & 2200\\
B3 & 500 & 10000 & 4000 & 200 & 100 & 15 & 60-120 & 2200\\
B4 & 500 & 12000 & 5000 & 200 & 100 & 15 & 60-120 & 2200\\
\addlinespace[0.8pt]
C1 & 100 & 8000 & 3000 & 200 & 100 & 15 & 60-120 & 2200\\
C2 & 200 & 8000 & 3000 & 200 & 100 & 15 & 60-120 & 2200\\
C3 & 300 & 8000 & 3000 & 200 & 100 & 15 & 60-120 & 2200\\
C4 & 400 & 8000 & 3000 & 200 & 100 & 15 & 60-120 & 2200\\
\addlinespace[0.8pt]
D1 & 500 & 8000 & 3000 & 200 & 100 & 5 & 60-120 & 2200\\
D2 & 500 & 8000 & 3000 & 200 & 100 & 10 & 60-120 & 2200\\
D3 & 500 & 8000 & 3000 & 200 & 100 & 20 & 60-120 & 2200\\
D4 & 500 & 8000 & 3000 & 200 & 100 & 30 & 60-120 & 2200\\
\addlinespace[0.8pt]
E1 & 100 & 500 & 300 & 200 & 100 & 5 & 60-120 & 2200\\
E2 & 100 & 500 & 300 & 300 & 150 & 5 & 60-120 & 3300\\
E3 & 100 & 500 & 300 & 400 & 200 & 5 & 60-120 & 4400\\
E4 & 200 & 500 & 300 & 400 & 200 & 5 & 60-120 & 4400\\
E5 & 200 & 750 & 400 & 400 & 200 & 5 & 60-120 & 4400\\
E6 & 200 & 1000 & 500 & 400 & 200 & 5 & 60-120 & 4400\\
\bottomrule
\end{tabular*}
\vspace{1mm}
\end{table}

\begin{table*}[htb]
\centering
\caption{Comparison of computational results for four algorithms}
\label{tab:core_case_metrics}
\scriptsize
\setlength{\tabcolsep}{2.0pt}
\renewcommand{\arraystretch}{1.05}
\resizebox{\textwidth}{!}{%
\begin{tabular}{l*{15}{l}}
\toprule
\multirow{2}{*}{Case} & \multicolumn{4}{c}{T3L-DS} & \multicolumn{4}{c}{SA} & \multicolumn{4}{c}{A-SeTVBRP} & \multicolumn{3}{c}{CNP}\\
\cmidrule(lr){2-5}\cmidrule(lr){6-9}\cmidrule(lr){10-13}\cmidrule(lr){14-16}
 & \multicolumn{1}{c}{\makecell{Emg.\\cov.}} & \multicolumn{1}{c}{\makecell{Routine\\ret.}} & \multicolumn{1}{c}{Pre-empt.} & \multicolumn{1}{c}{Resched.} & \multicolumn{1}{c}{\makecell{Emg.\\cov.}} & \multicolumn{1}{c}{\makecell{Routine\\ret.}} & \multicolumn{1}{c}{Pre-empt.} & \multicolumn{1}{c}{Resched.} & \multicolumn{1}{c}{\makecell{Emg.\\cov.}} & \multicolumn{1}{c}{\makecell{Routine\\ret.}} & \multicolumn{1}{c}{Pre-empt.} & \multicolumn{1}{c}{Resched.} & \multicolumn{1}{c}{\makecell{Emg.\\cov.}} & \multicolumn{1}{c}{\makecell{Routine\\ret.}} & \multicolumn{1}{c}{Pre-empt.}\\
\midrule
A1 & 79.036 $\pm$ 3.326 & 99.998 $\pm$ 0.001 & 0.005 $\pm$ 0.002 & 84.391 $\pm$ 8.209 & 83.692 $\pm$ 2.368 & 99.999 $\pm$ 0.001 & 82.003 $\pm$ 0.257 & 99.999 $\pm$ 0.001 & 77.838 $\pm$ 4.820 & 99.998 $\pm$ 0.001 & 0.005 $\pm$ 0.003 & 77.702 $\pm$ 7.085 & 70.371 $\pm$ 4.430 & 99.952 $\pm$ 0.003 & 0.047 $\pm$ 0.003 \\
A2 & 76.447 $\pm$ 1.574 & 99.996 $\pm$ 0.002 & 0.021 $\pm$ 0.008 & 87.871 $\pm$ 8.383 & 80.423 $\pm$ 1.647 & 99.999 $\pm$ 0.001 & 83.397 $\pm$ 0.266 & 99.998 $\pm$ 0.001 & 75.814 $\pm$ 1.958 & 99.994 $\pm$ 0.003 & 0.022 $\pm$ 0.007 & 77.321 $\pm$ 8.517 & 65.645 $\pm$ 0.993 & 99.887 $\pm$ 0.010 & 0.112 $\pm$ 0.010 \\
A3 & 77.675 $\pm$ 0.992 & 99.996 $\pm$ 0.003 & 0.039 $\pm$ 0.013 & 92.791 $\pm$ 6.722 & 82.431 $\pm$ 0.947 & 99.996 $\pm$ 0.001 & 83.878 $\pm$ 0.255 & 99.996 $\pm$ 0.001 & 76.309 $\pm$ 1.105 & 99.99 $\pm$ 0.003 & 0.045 $\pm$ 0.016 & 79.159 $\pm$ 5.254 & 66.956 $\pm$ 0.646 & 99.675 $\pm$ 0.023 & 0.324 $\pm$ 0.023 \\
A4 & 79.413 $\pm$ 1.459 & 99.995 $\pm$ 0.001 & 0.047 $\pm$ 0.009 & 89.856 $\pm$ 2.517 & 84.214 $\pm$ 1.396 & 99.993 $\pm$ 0.001 & 84.152 $\pm$ 0.274 & 99.992 $\pm$ 0.002 & 78.418 $\pm$ 1.428 & 99.985 $\pm$ 0.005 & 0.053 $\pm$ 0.017 & 74.115 $\pm$ 5.000 & 66.502 $\pm$ 1.036 & 99.572 $\pm$ 0.021 & 0.427 $\pm$ 0.021 \\
\addlinespace[1pt]
B1 & 84.669 $\pm$ 1.677 & 99.995 $\pm$ 0.001 & 0.040 $\pm$ 0.018 & 85.652 $\pm$ 8.367 & 88.560 $\pm$ 1.969 & 99.999 $\pm$ 0.001 & 88.139 $\pm$ 0.209 & 99.999 $\pm$ 0.001 & 83.034 $\pm$ 1.858 & 99.989 $\pm$ 0.004 & 0.035 $\pm$ 0.008 & 70.672 $\pm$ 7.464 & 72.280 $\pm$ 2.010 & 99.621 $\pm$ 0.014 & 0.378 $\pm$ 0.014 \\
B2 & 74.961 $\pm$ 1.784 & 99.996 $\pm$ 0.004 & 0.020 $\pm$ 0.011 & 82.808 $\pm$ 7.023 & 80.024 $\pm$ 2.417 & 99.997 $\pm$ 0.001 & 83.886 $\pm$ 0.217 & 99.996 $\pm$ 0.001 & 73.762 $\pm$ 1.748 & 99.994 $\pm$ 0.004 & 0.026 $\pm$ 0.014 & 74.124 $\pm$ 8.235 & 65.519 $\pm$ 1.530 & 99.772 $\pm$ 0.021 & 0.227 $\pm$ 0.021 \\
B3 & 74.762 $\pm$ 0.765 & 99.996 $\pm$ 0.002 & 0.022 $\pm$ 0.010 & 81.723 $\pm$ 7.547 & 80.070 $\pm$ 1.220 & 99.997 $\pm$ 0.001 & 82.154 $\pm$ 0.252 & 99.996 $\pm$ 0.001 & 73.793 $\pm$ 1.056 & 99.994 $\pm$ 0.002 & 0.025 $\pm$ 0.009 & 76.209 $\pm$ 6.174 & 64.290 $\pm$ 2.245 & 99.805 $\pm$ 0.006 & 0.194 $\pm$ 0.006 \\
B4 & 76.539 $\pm$ 2.380 & 99.995 $\pm$ 0.002 & 0.024 $\pm$ 0.008 & 81.138 $\pm$ 3.686 & 82.233 $\pm$ 2.077 & 99.998 $\pm$ 0.001 & 81.757 $\pm$ 0.128 & 99.997 $\pm$ 0.001 & 75.660 $\pm$ 1.990 & 99.993 $\pm$ 0.003 & 0.024 $\pm$ 0.011 & 73.896 $\pm$ 3.650 & 65.797 $\pm$ 1.033 & 99.820 $\pm$ 0.020 & 0.179 $\pm$ 0.020 \\
\addlinespace[1pt]
C1 & 50.512 $\pm$ 1.631 & 99.977 $\pm$ 0.009 & 0.099 $\pm$ 0.018 & 77.828 $\pm$ 5.763 & 62.094 $\pm$ 2.040 & 99.074 $\pm$ 0.023 & 58.740 $\pm$ 0.322 & 98.425 $\pm$ 0.035 & 50.198 $\pm$ 1.571 & 99.976 $\pm$ 0.007 & 0.098 $\pm$ 0.020 & 76.794 $\pm$ 4.165 & 43.762 $\pm$ 2.319 & 99.615 $\pm$ 0.028 & 0.384 $\pm$ 0.028 \\
C2 & 53.136 $\pm$ 2.641 & 99.989 $\pm$ 0.003 & 0.068 $\pm$ 0.023 & 83.966 $\pm$ 5.181 & 61.527 $\pm$ 2.547 & 99.844 $\pm$ 0.003 & 71.343 $\pm$ 0.394 & 99.782 $\pm$ 0.005 & 52.815 $\pm$ 2.490 & 99.988 $\pm$ 0.004 & 0.066 $\pm$ 0.021 & 82.769 $\pm$ 4.187 & 48.251 $\pm$ 3.064 & 99.709 $\pm$ 0.038 & 0.290 $\pm$ 0.038 \\
C3 & 67.945 $\pm$ 1.503 & 99.993 $\pm$ 0.003 & 0.042 $\pm$ 0.010 & 84.490 $\pm$ 6.031 & 74.003 $\pm$ 1.925 & 99.986 $\pm$ 0.001 & 78.409 $\pm$ 0.166 & 99.982 $\pm$ 0.001 & 67.780 $\pm$ 1.138 & 99.989 $\pm$ 0.003 & 0.044 $\pm$ 0.013 & 77.723 $\pm$ 4.737 & 59.791 $\pm$ 0.618 & 99.758 $\pm$ 0.025 & 0.241 $\pm$ 0.025 \\
C4 & 71.075 $\pm$ 1.024 & 99.995 $\pm$ 0.001 & 0.026 $\pm$ 0.006 & 80.932 $\pm$ 8.213 & 76.112 $\pm$ 1.226 & 99.995 $\pm$ 0.002 & 81.299 $\pm$ 0.342 & 99.994 $\pm$ 0.003 & 70.033 $\pm$ 0.924 & 99.993 $\pm$ 0.001 & 0.026 $\pm$ 0.005 & 74.633 $\pm$ 7.158 & 59.950 $\pm$ 2.210 & 99.800 $\pm$ 0.010 & 0.199 $\pm$ 0.010 \\
\addlinespace[1pt]
D1 & 79.670 $\pm$ 3.315 & 99.996 $\pm$ 0.002 & 0.035 $\pm$ 0.009 & 91.556 $\pm$ 5.348 & 85.007 $\pm$ 3.039 & 99.998 $\pm$ 0.001 & 82.133 $\pm$ 0.288 & 99.997 $\pm$ 0.001 & 77.536 $\pm$ 2.961 & 99.991 $\pm$ 0.003 & 0.037 $\pm$ 0.009 & 78.861 $\pm$ 5.321 & 65.240 $\pm$ 2.445 & 99.782 $\pm$ 0.014 & 0.217 $\pm$ 0.014 \\
D2 & 76.625 $\pm$ 3.406 & 99.996 $\pm$ 0.003 & 0.024 $\pm$ 0.011 & 85.829 $\pm$ 8.442 & 81.948 $\pm$ 2.892 & 99.997 $\pm$ 0.002 & 83.430 $\pm$ 0.318 & 99.996 $\pm$ 0.003 & 74.982 $\pm$ 2.927 & 99.994 $\pm$ 0.001 & 0.023 $\pm$ 0.010 & 75.710 $\pm$ 7.760 & 65.269 $\pm$ 2.500 & 99.796 $\pm$ 0.003 & 0.203 $\pm$ 0.003 \\
D3 & 77.634 $\pm$ 0.862 & 99.993 $\pm$ 0.003 & 0.030 $\pm$ 0.008 & 78.755 $\pm$ 6.350 & 82.134 $\pm$ 0.897 & 99.996 $\pm$ 0.002 & 84.043 $\pm$ 0.176 & 99.995 $\pm$ 0.003 & 76.247 $\pm$ 0.757 & 99.990 $\pm$ 0.002 & 0.030 $\pm$ 0.008 & 68.822 $\pm$ 8.165 & 66.888 $\pm$ 1.622 & 99.781 $\pm$ 0.008 & 0.218 $\pm$ 0.008 \\
D4 & 78.924 $\pm$ 0.780 & 99.996 $\pm$ 0.001 & 0.023 $\pm$ 0.006 & 86.307 $\pm$ 3.741 & 83.779 $\pm$ 0.801 & 99.996 $\pm$ 0.001 & 84.342 $\pm$ 0.144 & 99.995 $\pm$ 0.001 & 78.305 $\pm$ 0.822 & 99.992 $\pm$ 0.002 & 0.027 $\pm$ 0.006 & 73.154 $\pm$ 9.007 & 68.368 $\pm$ 1.570 & 99.776 $\pm$ 0.015 & 0.223 $\pm$ 0.015 \\
\addlinespace[1pt]
E1 & 79.957 $\pm$ 2.716 & 98.950 $\pm$ 0.320 & 6.777 $\pm$ 0.555 & 84.703 $\pm$ 3.616 & 85.342 $\pm$ 2.152 & 99.573 $\pm$ 0.294 & 71.399 $\pm$ 0.665 & 99.402 $\pm$ 0.413 & 75.518 $\pm$ 2.742 & 97.624 $\pm$ 0.297 & 7.142 $\pm$ 0.816 & 66.698 $\pm$ 2.614 & 66.277 $\pm$ 0.981 & 93.196 $\pm$ 0.338 & 6.803 $\pm$ 0.338 \\
E2 & 77.758 $\pm$ 1.801 & 98.664 $\pm$ 0.330 & 9.079 $\pm$ 0.729 & 85.341 $\pm$ 3.132 & 83.552 $\pm$ 1.339 & 99.491 $\pm$ 0.294 & 72.251 $\pm$ 1.096 & 99.298 $\pm$ 0.408 & 72.631 $\pm$ 1.853 & 97.090 $\pm$ 0.216 & 9.150 $\pm$ 0.647 & 68.189 $\pm$ 1.306 & 63.429 $\pm$ 1.298 & 91.411 $\pm$ 0.399 & 8.588 $\pm$ 0.399 \\
E3 & 77.522 $\pm$ 2.200 & 98.528 $\pm$ 0.530 & 10.854 $\pm$ 1.495 & 86.717 $\pm$ 3.180 & 83.161 $\pm$ 1.215 & 99.324 $\pm$ 0.427 & 72.292 $\pm$ 0.943 & 99.069 $\pm$ 0.584 & 72.373 $\pm$ 1.217 & 96.687 $\pm$ 0.628 & 11.048 $\pm$ 1.366 & 70.167 $\pm$ 2.371 & 63.057 $\pm$ 0.582 & 89.925 $\pm$ 1.103 & 10.074 $\pm$ 1.103 \\
E4 & 83.462 $\pm$ 3.234 & 99.082 $\pm$ 0.218 & 9.217 $\pm$ 0.900 & 90.034 $\pm$ 2.398 & 88.123 $\pm$ 2.796 & 99.888 $\pm$ 0.076 & 83.926 $\pm$ 0.987 & 99.868 $\pm$ 0.090 & 78.198 $\pm$ 2.931 & 97.916 $\pm$ 0.490 & 9.637 $\pm$ 0.571 & 78.366 $\pm$ 5.105 & 70.679 $\pm$ 2.236 & 91.108 $\pm$ 0.823 & 8.891 $\pm$ 0.823 \\
E5 & 83.291 $\pm$ 1.745 & 99.523 $\pm$ 0.152 & 6.647 $\pm$ 0.893 & 92.919 $\pm$ 1.689 & 88.938 $\pm$ 1.789 & 99.842 $\pm$ 0.096 & 82.031 $\pm$ 0.433 & 99.808 $\pm$ 0.117 & 78.424 $\pm$ 1.925 & 98.673 $\pm$ 0.180 & 6.727 $\pm$ 0.534 & 80.283 $\pm$ 2.095 & 71.197 $\pm$ 0.785 & 93.539 $\pm$ 0.590 & 6.460 $\pm$ 0.590 \\
E6 & 83.744 $\pm$ 0.410 & 99.257 $\pm$ 0.314 & 7.201 $\pm$ 0.430 & 89.720 $\pm$ 4.133 & 89.272 $\pm$ 0.444 & 99.866 $\pm$ 0.051 & 80.660 $\pm$ 0.397 & 99.834 $\pm$ 0.063 & 77.931 $\pm$ 1.003 & 98.068 $\pm$ 0.441 & 7.404 $\pm$ 0.347 & 74.035 $\pm$ 4.938 & 70.949 $\pm$ 1.703 & 93.083 $\pm$ 0.454 & 6.916 $\pm$ 0.454 \\
\bottomrule
\end{tabular}%
}
\vspace{1mm}
\parbox{\textwidth}{\scriptsize 
Note: All values are percentages. Each entry reports the mean $\pm$ standard deviation across ten problem instances generated from ten independent random seeds under the same configuration; all four methods are evaluated on the same instance (seed). 
Emg. cov., routine ret., pre-empt., and resched. denote emergency coverage, routine retention, routine pre-emption, and routine rescheduling, respectively.
CNP does not include a routine-rescheduling stage, so its rescheduling rate is zero in every case and is omitted from the table.}
\end{table*}

Table~\ref{tab:scenarios} organises 22 distinct scenarios into five groups covering emergency-demand scale, routine background load, constellation size, demand-arrival concentration, and conflict-enhanced balanced-load conditions.
Every configuration is evaluated with ten independent random seeds.

\subsection{Compared methods}

All methods use the same committed schedule, emergency waves, observation opportunities, and initial states.
The three comparative methods are:

\begin{itemize}
    \item \emph{SA}: a centralised simulated-annealing reference \cite{han2022simulated} that starts from a greedy solution and runs 1,000 iterations for each emergency wave with access to the current global instance.
    \item \emph{A-SeTVBRP}: an adapted selective time-variant better reply process based on \cite{yangDistributedSatellitesDynamic2025}, where satellites update feasible local schedules by asynchronous better replies and cluster heads exchange states through direct ISLs.
    \item \emph{CNP}: a conventional contract-net protocol \cite{liu2022Bottom} baseline that uses the same event-specific head-centred clusters as T3L-DS.
\end{itemize}

\subsection{Evaluation metrics}

Four metrics assess emergency service and routine-plan preservation. Let $\mathcal{G}^{E}$ be the unique emergency cells, and let $\mathcal{C}^{E}_m$ contain those observed by method $m$ within an applicable request window. 
Let $\mathcal{G}^{R}$ and $\mathcal{C}^{R}_m$ denote the routine cells in the common input schedule and the final schedule, respectively. 
Finally, $\mathcal{I}^{R}_m$ contains routine cells invalidated by emergency insertion, and $\mathcal{I}_m^{R,\mathrm{rsc}}=\mathcal{I}^{R}_m\cap\mathcal{C}^{R}_m$ contains those subsequently rescheduled.

\begin{equation}
\begin{aligned}
\mathrm{Cov}^{E}_m&=\frac{|\mathcal{C}^{E}_m|}{|\mathcal{G}^{E}|},&
\mathrm{Ret}^{R}_m&=\frac{|\mathcal{C}^{R}_m|}{|\mathcal{G}^{R}|},\\
\mathrm{Pre}^{R}_m&=\frac{|\mathcal{I}^{R}_m|}{|\mathcal{G}^{R}|},&
\mathrm{Rsc}^{R}_m&=\frac{|\mathcal{I}_m^{R,\mathrm{rsc}}|}{|\mathcal{I}^{R}_m|}.
\end{aligned}
\label{eq:evaluation_metrics}
\end{equation}

The four quantities are emergency coverage, routine retention, routine pre-emption, and routine rescheduling. A cell observed outside its emergency window receives no emergency credit. When  $\mathcal{I}^{R}_m=\emptyset$, $\mathrm{Rsc}^{R}_m$ is defined as zero.

\subsection{Experimental results and analysis}
\label{sec:results_analysis}

Table~\ref{tab:core_case_metrics} reports the means and standard deviations of the four core metrics for all five groups.
Across the 22 configurations, SA provides 5.5\% more emergency coverage than T3L-DS on average.
The centralised search can use global information to consider a wider range of insertions, but it pre-empts 79.8\% of the committed routine cells, compared with 2.3\% for T3L-DS.
SA subsequently reschedules most of those cells, so its mean routine retention is only 0.13 percentage points higher.

Within the distributed comparison, T3L-DS exceeds A-SeTVBRP and CNP in mean emergency coverage by 2.1 and 11.1\%, respectively.
It also reschedules 10.7\% more displaced routine cells than A-SeTVBRP, while CNP has no rescheduling stage.
These results are consistent with candidate-level marginal awards retaining feasible parts of overlapping bids and inter-cluster coordination providing further opportunities for unresolved demand.

The following analyses examine where these aggregate differences arise.
Groups A, C, and D compare performance under different emergency-demand scale, constellation size, and level of temporal concentration, respectively, while holding the other principal settings fixed.
Group E then considers successive changes in emergency demand, available satellites, and routine background under stronger conflicts.

\subsubsection{Emergency-demand scale}

\begin{figure}[!htb]
\centering
\includegraphics[width=0.98\columnwidth]{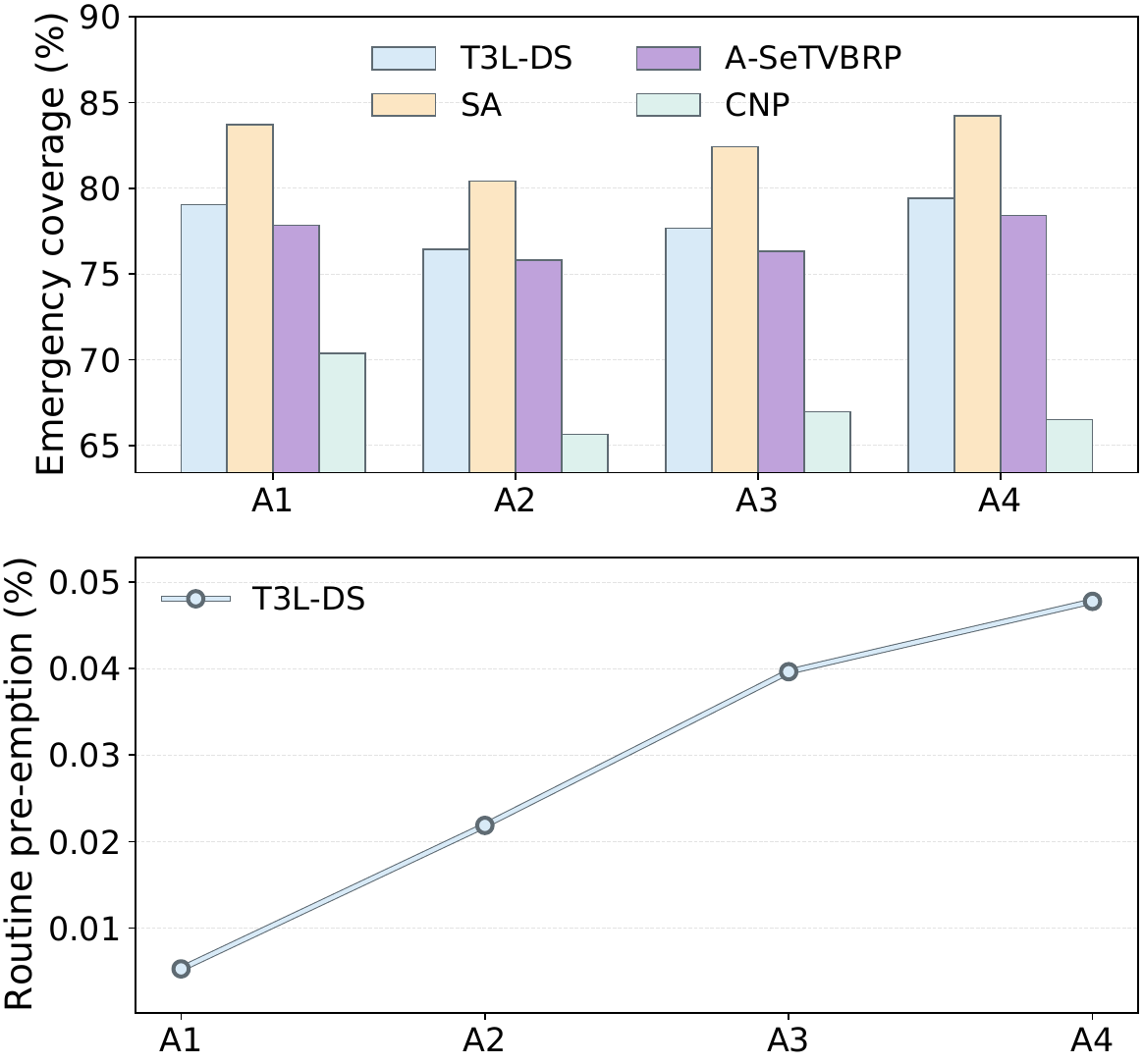}
\caption{Emergency coverage of the four methods (upper panel) and routine pre-emption of T3L-DS (lower panel) under increasing emergency demand. Cases A1-A4 keep the 500-satellite constellation and 11,000 routine targets fixed while increasing emergency requests from 75 to 600.}
\label{fig:result_group_A}
\end{figure}

Group A examines how emergency-demand scale affects performance while holding the constellation and routine background fixed.
Figure~\ref{fig:result_group_A} shows that emergency coverage initially decreases and then increases at higher demand levels across all four methods.
The initial growth in emergency demand intensifies competition for a fixed set of satellite access intervals, causing the covered proportion to fall.
As demand becomes denser, spatial overlap among emergency requests and between emergency and routine observations becomes more frequent.
A single observation can consequently satisfy several grid-cell demand units, which offsets part of the capacity pressure and causes coverage to rise at the higher demand levels.

The difference between T3L-DS and SA stays from $4.0\%$ to $4.8\%$.
T3L-DS remains ahead of A-SeTVBRP and establishes a larger advantage over CNP at the highest demand level.
From A2 to A4, the emergency coverage of T3L-DS increases by $3.9\%$, compared with $3.4\%$ for A-SeTVBRP and $1.3\%$ for CNP.
Candidate-level awards retain useful observations from overlapping bids, and inter-cluster coordination reallocates demand that cannot be accommodated within its initial cluster.
These mechanisms become more valuable as simultaneous conflicts increase.
Routine pre-emption also rises with emergency demand because more insertions encounter occupied intervals, although the increase becomes slower at the highest demand level.
At this density, same-cell reuse serves more additional demand without requiring a proportional increase in displaced routine observations.

\subsubsection{Constellation size}

\begin{figure}[!htb]
\centering
\includegraphics[width=0.98\columnwidth]{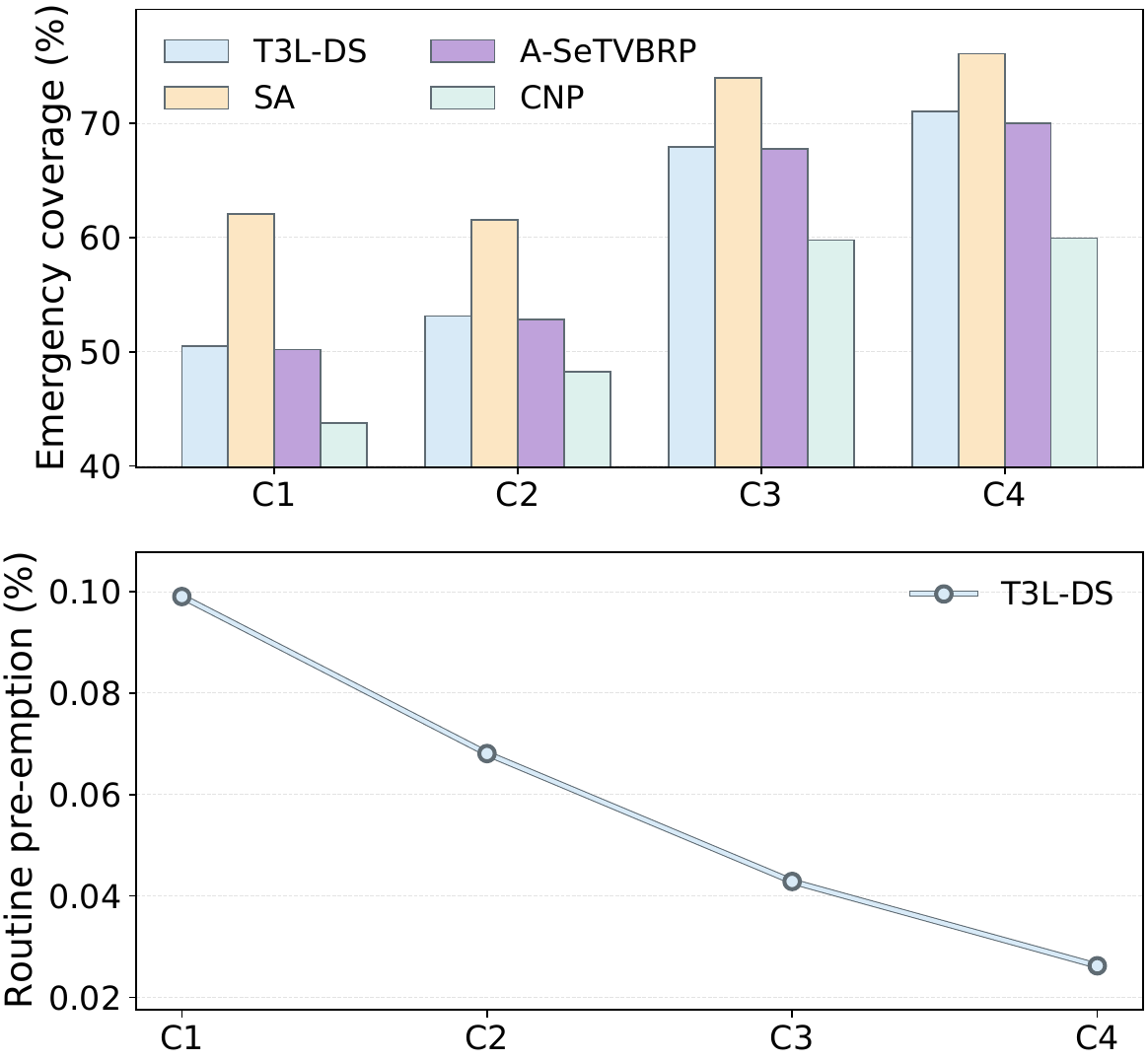}
\caption{Emergency coverage of the four methods (upper panel) and routine pre-emption of T3L-DS (lower panel) under increasing constellation size. Cases C1-C4 keep the routine and emergency demand fixed while increasing the constellation from 100 to 400 satellites.}
\label{fig:result_group_C}
\end{figure}

Group C examines how constellation size affects performance while keeping routine and emergency demand unchanged.
Figure~\ref{fig:result_group_C} shows a pronounced rise in emergency coverage as more satellites become available.
From C1 to C4, T3L-DS increases its emergency coverage by $40.7\%$, compared with $22.6\%$ for SA.
The relative gain is larger for T3L-DS because constellation expansion supplies more feasible local candidates and cross-cluster alternatives, reducing the disadvantage of making decisions with local information.
SA already exploits constellation-wide opportunities in the smaller cases, so its relative benefit from the added satellites is lower.
For all methods, the improvement becomes less pronounced at the largest constellation size because new satellites increasingly duplicate visibility that is already available.

T3L-DS remains slightly ahead of A-SeTVBRP across the group and reaches a larger advantage over CNP at the higher constellation sizes.
Inter-cluster coordination and candidate-level evaluation allow T3L-DS to use the enlarged opportunity set without accepting or rejecting each bid as a whole.
Routine pre-emption falls by $73.7\%$ from C1 to C4 because more emergency observations can be assigned to alternative satellites and idle intervals.
The reduction gradually flattens as the remaining conflicts concentrate on requests with limited visibility or restrictive time windows.

\subsubsection{Emergency-demand temporal concentration}

\begin{figure}[!htb]
\centering
\includegraphics[width=0.98\columnwidth]{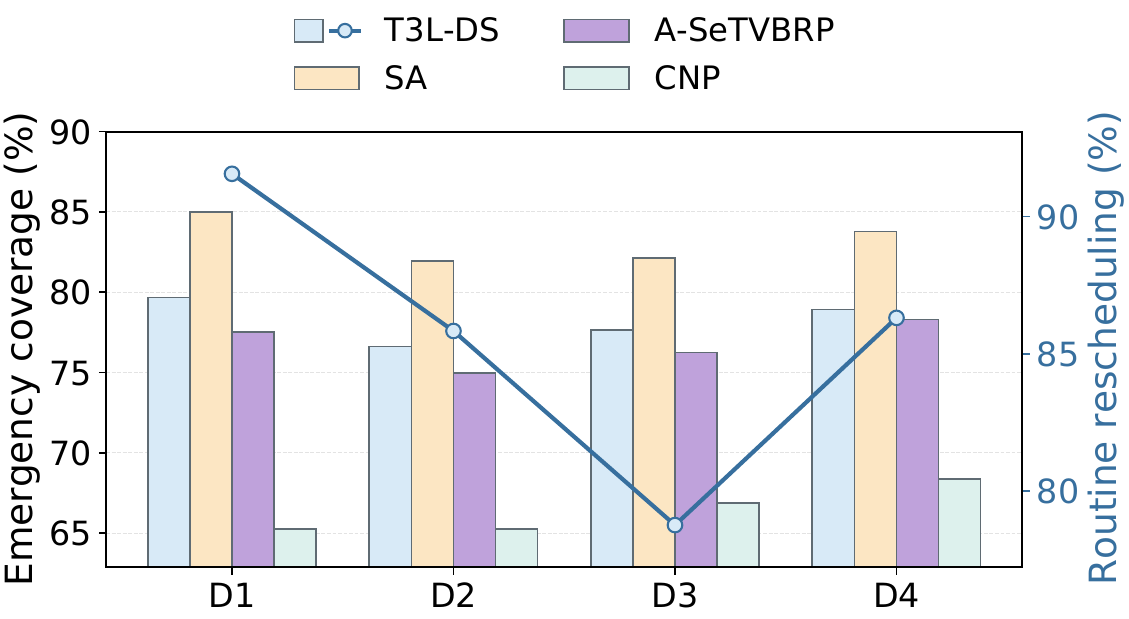}
\caption{Effect of the emergency arrival-wave count on coverage and routine rescheduling. Bars show the emergency coverage of the four methods on the left axis, and the line shows the routine rescheduling of T3L-DS on the right axis. Cases D1-D4 keep 300 emergency requests fixed while distributing them over 5 to 30 arrival waves.}
\label{fig:result_group_D}
\end{figure}

Group D keeps total emergency demand fixed and changes the number of arrival waves.
Figure~\ref{fig:result_group_D} shows that coverage first decreases and then increases as the requests are divided into more waves.
With five waves, each allocation event contains more requests and provides greater scope for shared grid coverage and coordinated assignment.
Increasing the number of waves initially separates requests that could otherwise be considered together, while earlier insertions occupy opportunities available to later waves.
At the highest wave count, the number of simultaneous requests becomes sufficiently small to reduce competition within each allocation event, and coverage increases.

T3L-DS remains the strongest distributed method, while its difference from SA ranges from $4.5\%$ to $5.3\%$.
Its advantage over A-SeTVBRP narrows as each wave becomes smaller because fewer requests require cross-satellite alternatives at the same time.
CNP also benefits from lower per-wave competition, but whole-bid allocation continues to lose useful observations when bids overlap.
Routine rescheduling also decreases at first and then increases.
At intermediate wave counts, repeated emergency commitments divide the remaining schedule into shorter feasible intervals.
The smaller request set in the D4 event permits more selective insertions and leaves more feasible gaps, allowing the T3L-DS rescheduling rate to increase by $9.6\%$ relative to D3.

\subsubsection{Conflict-enhanced balanced-load conditions}

\begin{figure}[!htb]
\centering
\includegraphics[width=0.98\columnwidth]{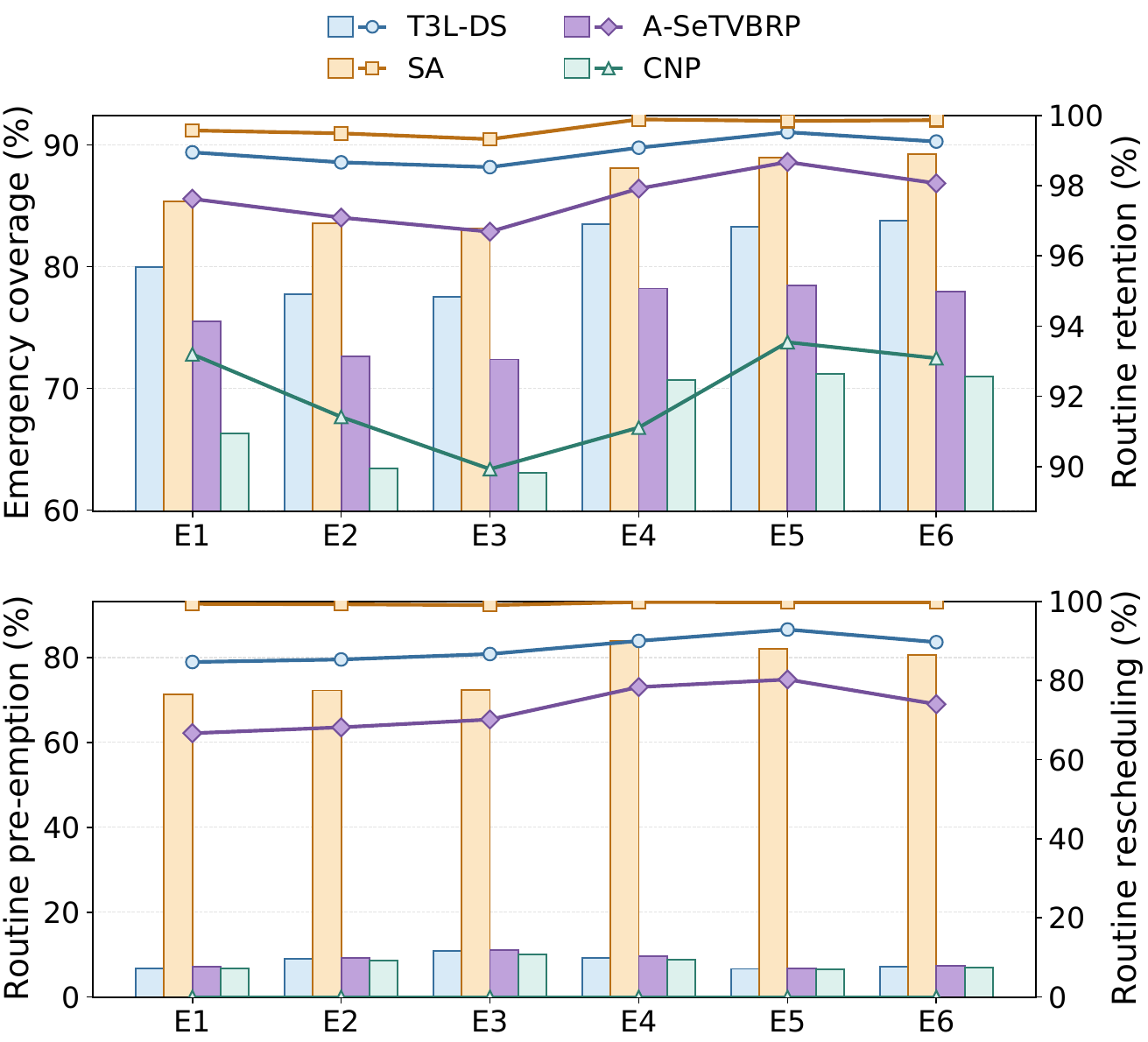}
\caption{Performance under conflict-enhanced balanced-load conditions. The upper panel combines emergency coverage (bars, left axis) and routine retention (lines, right axis); the lower panel combines routine pre-emption (bars, left axis) and routine rescheduling (lines, right axis). 
Cases E1-E3 use 100 satellites with a fixed routine load and increasing emergency demand. Cases E3-E4 keep both loads fixed and increase the constellation from 100 to 200 satellites. Cases E4-E6 use 200 satellites with fixed emergency demand and increasing routine load.
}
\label{fig:result_group_E}
\end{figure}

Group E contains three controlled comparisons under stronger spatial and temporal conflicts.
Each comparison changes only one of emergency demand, constellation size, and routine background while holding the other settings fixed.
Figure~\ref{fig:result_group_E} reports emergency service and routine-plan preservation across these comparisons.
E1-E3 increase emergency demand while holding the 100-satellite constellation and routine load fixed.
Coverage declines for all four methods, but T3L-DS loses $3.1\%$ relative to E1, compared with $4.2\%$ for A-SeTVBRP and $4.9\%$ for CNP.
Its difference from SA remains almost unchanged, while its advantage over the distributed baselines grows as conflicts become denser.
Partial bid retention avoids discarding non-conflicting observations, and inter-cluster coordination broadens the set of satellites that can absorb unresolved demand.
Routine retention declines and pre-emption rises over the same cases because the additional emergency insertions consume a larger share of the adjustable routine plan.

E3-E4 compare performance before and after the constellation size is doubled.
T3L-DS improves its emergency coverage by $7.7\%$ relative to E3, exceeding the $6.0\%$ increase of SA while retaining its advantage over the distributed baselines.
Under the stronger conflicts in Group E, the added satellites supply alternative access intervals and reduce the number of requests confined to a small set of feasible spacecraft.
T3L-DS can exploit these alternatives through inter-cluster coordination, whereas SA already uses global information and obtains a smaller relative gain from the additional resources.
Routine retention rises at the same time and routine pre-emption falls.
The additional emergency coverage is obtained from new orbital opportunities rather than from greater disruption of the routine plan.

E4-E6 then increase only the routine background.
Relative to E4, T3L-DS changes its emergency coverage by no more than $0.4\%$ in E5 and E6, so the added routine demand mainly affects the routine plan.
The pre-emption rate remains below its E4 level because displaced grid cells form a smaller share of the expanded routine plan and more emergency cells can reuse scheduled observations.
Routine rescheduling also stays close to its E4 level despite the denser background.
Together, the two panels show that T3L-DS preserves emergency service under severe conflict and reschedules most of the displaced routine cells.

\subsection{Ablation study}

\begin{figure}[!htb]
\centering
\includegraphics[width=0.98\columnwidth]{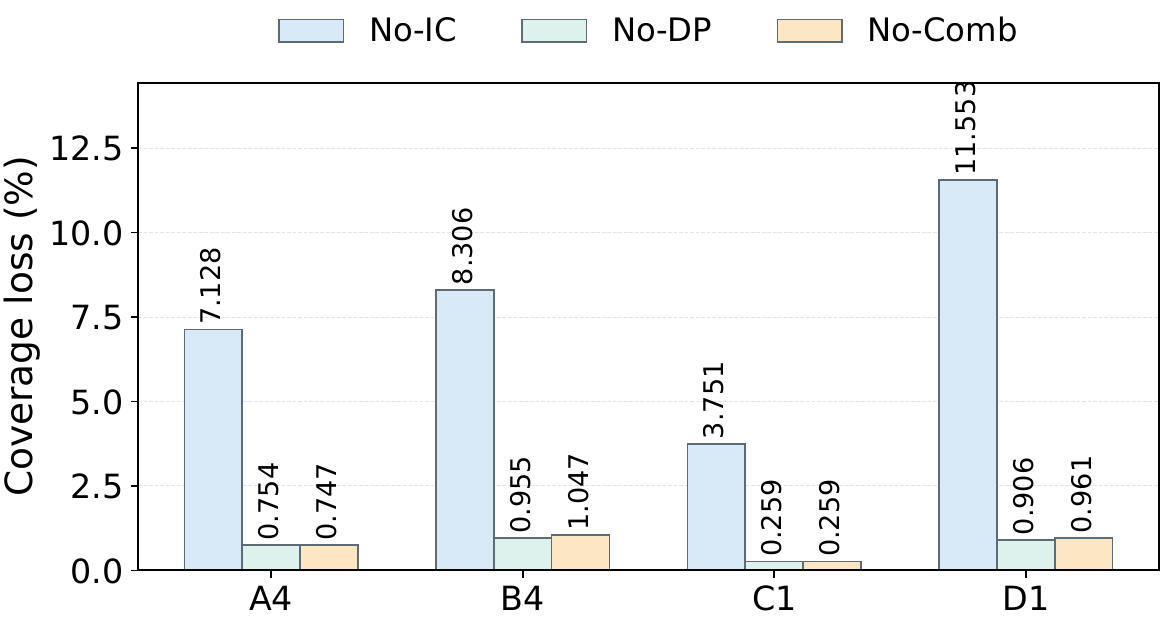}
\caption{Emergency-coverage loss relative to Full under the three ablation settings.
A4, B4, C1, and D1 represent emergency-demand, routine-load, constellation-size, and temporal-concentration stress, respectively.}
\label{fig:ablation_coverage_loss}
\end{figure}

T3L-DS comprises three principal coordination mechanisms: inter-cluster coordination, dual-plan construction, and candidate-level joint evaluation.
An ablation study is conducted to quantify the contribution of each mechanism to the overall scheduling performance.
The three reduced settings remove inter-cluster coordination (No-IC), retain Plan~A only (No-DP), and combine Plan-A-only generation with whole-plan acceptance in place of candidate-level joint evaluation (No-Comb), respectively.
It uses A4 and B4, which impose the largest emergency and routine loads in their respective groups, C1 with the smallest constellation, and D1 with the most concentrated arrivals.
Table~\ref{tab:ablation_core_metrics} reports the four core metrics, while Figure~\ref{fig:ablation_coverage_loss} shows the loss of emergency coverage relative to the full method.

\begin{table}[htb]
\centering
\caption{Ablation results}
\label{tab:ablation_core_metrics}
\fontsize{6.2pt}{7.2pt}\selectfont
\setlength{\tabcolsep}{1.0pt}
\renewcommand{\arraystretch}{1.00}
\begin{tabular*}{\columnwidth}{@{\extracolsep{\fill}}cl*{4}{l}@{}}
\toprule
Case & Variant & \multicolumn{1}{c}{Emg. cov.} & \multicolumn{1}{c}{Routine ret.} & \multicolumn{1}{c}{Pre-empt.} & \multicolumn{1}{c}{Resched.}\\
\midrule
\multirow{4}{*}{A4} & Full & 79.413 $\pm$ 1.459 & 99.995 $\pm$ 0.001 & 0.047 $\pm$ 0.009 & 89.856 $\pm$ 2.517\\
 & No-IC & 72.285 $\pm$ 1.056 & 99.989 $\pm$ 0.004 & 0.044 $\pm$ 0.012 & 76.054 $\pm$ 3.367\\
 & No-DP & 78.659 $\pm$ 1.559 & 99.996 $\pm$ 0.001 & 0.044 $\pm$ 0.010 & 90.997 $\pm$ 2.426\\
 & No-Comb & 78.665 $\pm$ 1.746 & 99.996 $\pm$ 0.001 & 0.045 $\pm$ 0.007 & 90.948 $\pm$ 2.362\\
\addlinespace[1pt]
\multirow{4}{*}{B4} & Full & 76.539 $\pm$ 2.380 & 99.995 $\pm$ 0.002 & 0.024 $\pm$ 0.008 & 81.138 $\pm$ 3.686\\
 & No-IC & 68.232 $\pm$ 2.924 & 99.993 $\pm$ 0.003 & 0.021 $\pm$ 0.010 & 70.275 $\pm$ 3.588\\
 & No-DP & 75.583 $\pm$ 2.325 & 99.995 $\pm$ 0.002 & 0.023 $\pm$ 0.008 & 80.752 $\pm$ 4.750\\
 & No-Comb & 75.492 $\pm$ 2.270 & 99.995 $\pm$ 0.002 & 0.022 $\pm$ 0.009 & 79.861 $\pm$ 4.098\\
\addlinespace[1pt]
\multirow{4}{*}{C1} & Full & 50.512 $\pm$ 1.631 & 99.977 $\pm$ 0.009 & 0.099 $\pm$ 0.018 & 77.828 $\pm$ 5.763\\
 & No-IC & 46.761 $\pm$ 1.264 & 99.978 $\pm$ 0.007 & 0.088 $\pm$ 0.016 & 75.670 $\pm$ 4.016\\
 & No-DP & 50.252 $\pm$ 1.583 & 99.977 $\pm$ 0.009 & 0.096 $\pm$ 0.021 & 77.569 $\pm$ 5.413\\
 & No-Comb & 50.252 $\pm$ 1.830 & 99.977 $\pm$ 0.009 & 0.075 $\pm$ 0.018 & 77.691 $\pm$ 5.153\\
\addlinespace[1pt]
\multirow{4}{*}{D1} & Full & 79.670 $\pm$ 3.315 & 99.996 $\pm$ 0.002 & 0.035 $\pm$ 0.009 & 91.556 $\pm$ 5.348\\
 & No-IC & 68.117 $\pm$ 2.731 & 99.992 $\pm$ 0.003 & 0.032 $\pm$ 0.007 & 77.797 $\pm$ 5.246\\
 & No-DP & 78.765 $\pm$ 3.207 & 99.996 $\pm$ 0.002 & 0.034 $\pm$ 0.007 & 91.061 $\pm$ 6.546\\
 & No-Comb & 78.709 $\pm$ 3.046 & 99.996 $\pm$ 0.002 & 0.035 $\pm$ 0.006 & 90.465 $\pm$ 6.269\\
\bottomrule
\end{tabular*}
\vspace{1mm}
\parbox{\columnwidth}{\scriptsize All values are percentages. Each entry reports the mean $\pm$ standard deviation across ten problem instances generated from ten independent random seeds under the same configuration.  
Emg. cov., routine ret., pre-empt., and resched. denote emergency coverage, routine retention, routine pre-emption, and routine rescheduling, respectively.
}
\end{table}

As shown in Table~\ref{tab:ablation_core_metrics} and Figure~\ref{fig:ablation_coverage_loss}, inter-cluster coordination makes the largest contribution among the three components evaluated.
Its removal reduces emergency coverage by $3.8\%$-$11.6\%$ across the four selected cases.
The largest loss occurs in D1, where each arrival wave contains more simultaneous requests, followed by B4 with its dense committed routine plan and A4 with its heavy emergency load.
In these cases, the visibility and attitude margins available inside one temporary cluster are more readily exhausted, so neighbouring clusters provide useful alternatives.
The smaller loss in C1 is consistent with its sparse 100-satellite topology, which offers fewer neighbouring assets capable of accepting transferred demand.

Local-plan diversity and candidate-level joint evaluation have smaller effects on coverage.
Removing Plan B produces losses of $0.26\%$-$0.96\%$, with the larger changes in B4 and D1.
Under a dense routine plan or concentrated arrivals, a second local plan can retain an alternative that avoids a conflict affecting Plan A.
Replacing candidate-level joint evaluation with whole-plan acceptance changes coverage by no more than $0.09\%$ beyond the Plan-A-only setting.
The results therefore retain the original ordering of contributions.
Inter-cluster coordination provides the main gain under resource pressure, while dual-plan construction and candidate-level evaluation refine the local choices.

\section{Conclusions}
\label{sec:conclusion}

This paper studies DEOSP, where emergency point and area requests must be scheduled while a routine observation plan is already being executed. 
The proposed geographic-grid formulation uses common spatial units to describe emergency demand, satellite footprints, coverage credit, and disturbance to the committed plan. 
On this basis, T3L-DS coordinates scheduling through event-level ground organisation, onboard dual-plan bidding, and cluster-level allocation over available ISLs. 
The method modifies only adjustable future activities, while completed, executing, and protected routine activities remain outside the revision scope.

The experiments show that T3L-DS gives the strongest emergency service among the distributed methods and preserves most routine grid-cell coverage under the tested conditions. 
Future work will add more detailed onboard computation, energy, and communication models, and will test the framework under uncertain ISL availability.

\section*{Acknowledgements}
This work was supported by the National Natural Science Foundation of China under Grants 62503503 and 62373380.
The authors used a large language model to assist with language editing, LaTeX consistency checking, and manuscript organisation.
All content was reviewed by the authors, who take full responsibility for the manuscript.

\bibliographystyle{unsrt}
\bibliography{area_target}

\end{document}